\documentclass{article} 
\usepackage[numbers]{natbib}

\usepackage{amsmath,amsfonts,bm}

\def\eqref#1{equation~\ref{#1}}

\def\1{\bm{1}}

\DeclareMathAlphabet{\mathsfit}{\encodingdefault}{\sfdefault}{m}{sl}
\SetMathAlphabet{\mathsfit}{bold}{\encodingdefault}{\sfdefault}{bx}{n}

\usepackage{hyperref}
\usepackage{url}
\usepackage{multirow}
\usepackage{wrapfig}

\makeatletter
\newcommand{\specificthanks}[1]{\@fnsymbol{#1}}
\makeatother

\title{Out-of-distribution Prediction with Invariant Risk Minimization: The Limitation and An Effective Fix}

\author{\begin{tabular}{cc}Ruocheng Guo \thanks{Part of the work is done during the authors' internship at Microsoft Research, Redmond} & Pengchuan Zhang \\
Arizona State University & Microsoft Research\\
rguo12@asu.edu & penzhan@microsoft.com\\ \\
Hao Liu\textsuperscript{ \specificthanks{1}} & Emre Kiciman\\
Caltech & Microsoft Research\\
hliu3@caltech.edu & emrek@microsoft.com\\
\end{tabular}}

\usepackage{tikz-qtree}
\usepackage{tikz-qtree-compat}
\usepackage{dsfont}
\usepackage{amsmath}
\usepackage{graphicx}
\usepackage{subfigure}
\usepackage{graphics}

\usepackage{wrapfig}
\usepackage{color}

\newcommand{\indep}{\perp \! \! \! \perp} 

\usetikzlibrary{shapes,decorations,arrows,calc,arrows.meta,fit,positioning}
\tikzset{
	-Latex,auto,node distance =1 cm and 1 cm,semithick,
	state/.style ={ellipse, draw, minimum width = 0.7 cm},
	point/.style = {circle, draw, inner sep=0.04cm,fill,node contents={}},
	bidirected/.style={Latex-Latex,dashed},
	el/.style = {inner sep=2pt, align=left, sloped}
}
\begin{document}
\date{}

\maketitle

\begin{abstract}
This work considers out-of-distribution (OOD) prediction with training data from multiple domains and test data from a novel domain. DNNs fail in OOD prediction because they are prone to pick up spurious correlations. Invariant Risk Minimization (IRM) is proposed to address this issue. Its effectiveness has been shown in the colored MNIST experiment. Nevertheless, IRM only guarantees the existence of an invariant optimal classifier for those overlapping feature representations across domains. As DNNs tend to learn shortcuts, they can circumvent IRM by learning non-overlapping representations for different domains. To show this, we consider a setting, \emph{strong triangle spuriousness} -- when the spurious correlations among spurious features, the domain variable, and the class label are stronger than the invariant correlation between invariant features and the class label. In this setting, DNNs can learn non-overlapping feature representations and achieve low empirical risk at the same time. Empirically, we show that the performance of IRM can be dramatically degraded under this setting. In this work, we try to answer the questions: why does IRM fail in the aforementioned setting? Why does IRM work for the original colored MNIST dataset? With a series of semi synthetic datasets -- the colored MNIST plus, we expose the problem of IRM and demonstrates the efficacy of the proposed method.
\end{abstract}

\section{Introduction}






Strong empirical results have demonstrated the efficacy of deep neural networks (DNNs) in various areas including computer vision, natural language processing and speech recognition.
However, such positive results overwhelmingly rely on the assumption that the training and test data are independent identical samples of the same underlying distribution.
In contrast, in out-of-distribution (OOD) prediction, we consider training data from multiple domains and test data from an unseen distribution from training, the performance of DNNs can be dramatically degraded.
This is because DNNs are prone to pick up spurious correlations which do not hold beyond the training dataset~\citep{beery2018recognition,arjovsky2019invariant,geirhos2020shortcut}.
For example, when most camel pictures in a training set have a desert in the background, DNNs will pick up the spurious correlation between desert and camel, leading to failures when camel pictures come with different backgrounds in the test set.
Therefore, OOD prediction remains a challenging problem for DNNs.

%

The invariant correlations across different domains turn out to be the key to address the challenge of OOD prediction.
%
%
Spurious correlations learned in one domain are unreliable in another, the invariant ones enable DNNs to generalize to unseen domains.
In practice, it is extremely difficult to know whether an input feature has invariant or spurious correlation with the label.
Thus, Invariant Risk Minimization (IRM)~\citep{arjovsky2019invariant} proposes a recipe for training DNNs to capture invariant correlations which generalize to unseen domains.
A DNN model consists of a representation learning module and a predictor.
IRM captures invariant correlations by learning representations that elicit an optimal invariant predictor across training domains.
When the invariant relationship between representations and the label is linear, IRM can be reduced to its practical form, a regularizer that can be optimized along with empirical risk by stochastic gradient descent~\citep{arjovsky2019invariant} .
%
%
%
%
However, \cite{arjovsky2019invariant} mentioned that IRM would only penalize DNNs when the representations that pick up spurious correlations overlap across domains.
DNNs tend to take shortcuts~\citep{geirhos2020shortcut}.
So, they can circumvent the regularization of IRM with non-overlapping representations for different domains.
There exist multiple invariant optimal classifiers elicited by non-overlapping representations that lead to dramatically different performance in unseen domains, which is one of the reasons to explain the phenomenon of model underspecification~\citep{d2020underspecification}.
%

%
%
%


\begin{figure}[tb]
    \begin{minipage}{0.49\textwidth}
        \centering
      \small
      \subfigure[Training]{\label{subfig:causal_graph_training}\begin{tikzpicture}
\centering

    \node[state] (e) at (3,2) {$E$};
    \node[state] (c) at (3,0) {$C$};
    
     \node[state] (x) at (-1,0) {$X$};
        \node[state] (y) at (1,1) {$Y$};
        
    \node[state] (ys) at (1,2) {$Y^*$};
    \node[state] (s) at (-1,2) {$S$}; 

    \path [black,line width=1.0] (e) edge (c);

    \path [black,line width=1.0] (y) edge (c);
               \path [black,line width=1.0] (y) edge (e);
        \path [black,line width=1.0] (c) edge (x);
            \path [black,line width=1.0] (s) edge (x);
            \path [black,line width=1.0] (ys) edge (y);
                \path [black,line width=1.0] (s) edge (ys);

\end{tikzpicture}}
    \end{minipage}
\begin{minipage}{0.49\textwidth}
        \centering
      \small
      \subfigure[Test]{\label{subfig:causal_graph_test}\begin{tikzpicture}
\centering

    \node[state] (e) at (3,2) {$E$};
    \node[state] (c) at (3,0) {$C$};
    
     \node[state] (x) at (-1,0) {$X$};
        \node[state] (y) at (1,1) {$Y$};
        
    \node[state] (ys) at (1,2) {$Y^*$};
    \node[state] (s) at (-1,2) {$S$}; 

    \path [black,line width=1.0] (e) edge (c);

    \path [black,line width=1.0] (y) edge (c);
        \path [black,line width=1.0] (c) edge (x);
            \path [black,line width=1.0] (s) edge (x);
            \path [black,line width=1.0] (ys) edge (y);
                \path [black,line width=1.0] (s) edge (ys);

\end{tikzpicture}}
    \end{minipage}
    \caption{
    The causal graphs of the Colored MNIST Plus dataset (CMNIST+) used in this work. They mimic the data generating process of the original Colored MNIST (CMNIST)~\citep{arjovsky2019invariant,ahuja2020empirical}.
    The observed features $X$ are determined by the shape $S$ and the color $C$.
    The observed label $Y$ is generated by randomly flipping the true label $Y^*$, which is decided by the shape $S$ (the invariant feature).
    The color $C$ is generated based on the domain $E$ and the observed label $Y$.
    $P(Y|S)$ is the invariant relationship across domains, while the spurious correlation $P(Y|C)$ varies. 
        This is because, when there is an intervention on $E$, $P(E|Y)$ and $P(C|Y,E)$ would vary.
This implies that the relationships among $E$, $Y$ and $C$ are spurious, which is called \emph{triangle spuriousness} in this work.
    }
    \label{fig:causal_graph_cmnist_plus}
    \end{figure}
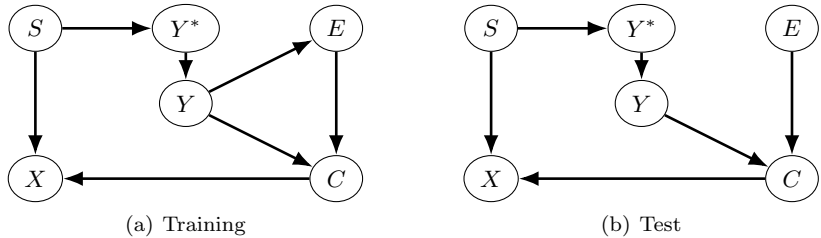

In this work, we consider the setting where DNNs can learn non-overlapping spurious representations to circumvent the regularization of IRM and minimize the empirical risk at the same time.
In particular, we let the training data come with spurious correlations among the spurious features, the class label and the domain variable.
%
%
When the spurious correlations are stronger than the invariant relationship, we name this setting \emph{strong triangle spuriousness}. In this setting, IRM regularized empirical risk can take low values when DNNs learn spurious and non-overlapping representations that accurately predict the domain label.
%
%
This is because, in this setting, picking up such spurious features can achieve high accuracy in predicting both domain and class in the training domains, but not in the test.
However, the original colored MNIST dataset cannot expose this issue as the strong similarity between the two training domains makes it unlikely for DNNs to pick up the weak spurious correlation between the domain label and the spurious features.
To empirically illustrate this problem, we design a new dataset -- the colored MNIST plus (CMNIST+).
Fig.~\ref{fig:causal_graph_cmnist_plus} shows the causal graph of CMNIST+.
A valid intervention on the domain variable $E$ would make the spurious relationships among $Y$, $E$ and $C$ vary across domains\footnote{An intervention is valid iff it does not influence the value of the label $Y$~\citep{ahuja2020empirical}.}.
As shown in Fig.~\ref{fig:irm_cmnist_plus}, in this dataset, when triangle spuriousness is stronger than the invariant relationship ($\rho > 0.75$), the performance of IRM models significantly degrades.
%

Moreover, to resolve this issue of IRM, we propose an effective solution, which combines IRM with conditional distribution matching (CDM).
The CDM regularization pushes the representation distribution of instances from the same class to be overlapping across domains.
We show that (1) a combination of CDM and IRM regularization helps DNNs learn invariant representations and (2) CDM can prevent DNNs from circumventing the IRM regularization with non-overlapping representations across domains.
Empirically, on our newly introduced dataset, the proposed method achieves significant performance improvement over IRM under strong triangle spuriousness.

%

%
\section{Related Work}

In this section, we briefly review the literature of OOD prediction and domain adaptation.

%
%
%
IRM \citep{arjovsky2019invariant} formulates causal feature learning as a constraint on the ERM framework~\citep{vapnik1992principles}, which imposes the causal inductive bias: causal feature representations lead to the existence of an optimal classifier for all training domains.
Then it is transformed to a regularizer which can be minimized along with empirical risks.
The most relevant work to ours includes~\citep{rosenfeld2020risks} and~\citep{ahuja2020empirical}.
\cite{rosenfeld2020risks} theoretically show that, when features are generated by a function of invariant and spurious independent Gaussian factors which have linear correlations with the label, IRM would learn spurious representations when the dimensionality of invariant features is greater than the number of domains.
Note that the higher the dimensionality is, the less likely representations across domains would overlap.
\cite{ahuja2020empirical} argue that IRM only outperforms ERM when representations of different domains significantly overlap.
These observations can explain why ERM performs better than IRM in the DOMAINBED benchmark~\citep{gulrajani2020search} as high-dimensional representations are used in their experiments.
%
%
%
\cite{ahuja2020invariant} reformulate the optimization problem of IRM from a game theory aspect.
%
%
Naturally, OOD prediction also has a robust optimization formulation~\citep{ben2009robust}, which aims to minimizes the worst empirical risk across training domains.
%
%
%
\cite{krueger2020out} extend robust optimization to minimize the empirical risk of the worst domain and maximize that of other domains.
They also propose to minimize the variance of domain specific empirical risks along with the empirical risk.
\cite{jin2020domain} propose to minimize the domain specific risk between two models, one trained on the same domain, the other trained on the other domains.
These methods essentially minimize the differences among domain specific risks.
%
%
Data augmentation can also improve the generalizability of DNNs~\citep{ilse2020designing}.
%
%
%
%
%
%
%
However, augmentation requires prior knowledge on the differences among domains, which may not be available in OOD prediction.
%
%
\cite{zhang2020causal} propose to make DNNs more robust against test data generated by unseen interventions.
They model interventions in training data with a generative model and perform test-time inference to catch unseen interventions.
Compared to the existing work, this work focuses on the undesired solutions of IRM in the setting of strong triangle spuriousness. We design a dataset to empirically expose this problem of IRM and propose a simple but effective fix to it.
%

%

Domain Adaptation (DA) assumes that the unlabeled test set can be used during training and validation.
%
%
From the methodology aspect, distribution matching methods used in DA, such as gradient reversing~\citep{ganin2015unsupervised} and adversarial CDM~\citep{long2018conditional} are useful for OOD prediction.
\cite{peters2016causal} realize the invariance of causal relationships can be used for DA.
\cite{zhang2013domain} propose reweighting and kernel based distribution matching methods to handle three types of DA problems: target shift, conditional shift and generalized target shift.
\cite{gong2016domain} work on extracting transferable components $F(X)$ that ensure $P(F(X)|Y)$ to be invariant across domains with location-scale transformation. Their method can identify how $P(Y)$ changes across domains simultaneously.
Different from DA, we strictly ensure that the test domain is unseen during training and validation to reflect the scenario of OOD prediction in real-world applications.

\section{Preliminaries and IRM}

%



\noindent\textbf{Notations.}
We use lowercase (e.g., $x$), uppercase (e.g., $X$) and calligraphic uppercase (e.g., $\mathcal{X}$) letters for values, random variables and spaces.
We let $X \in \mathcal{X}$, $Y\in \mathcal{Y}$ and $E \in \mathcal{E}$ denote raw input, the class label and the domain label where $\mathcal{X}$, $\mathcal{Y}$ and $\mathcal{E}$ are the spaces of input, class labels and domains.
A DNN model consists of a feature learning function $F$ and a classifier $G$.
A feature learning function $F:\mathcal{X}\rightarrow \mathds{R}^d$ maps raw input $X$ to its $d$-dimensional representations $F(X)$.
A classifier $G:\mathds{R}^d\rightarrow \mathcal{Y}$ maps a feature representation to a class label.
We denote their parameters by $\bm{\theta}_F$ and $\bm{\theta}_G$, respectively.
Let $\bm{\theta}=Concat(\bm{\theta}_F,\bm{\theta}_G)$ denote the concatenation of them. 

A domain $e$ of $n_e$ instances is denoted by $D_e=\{x_i^e,y_i^e\}_{i=1}^{n_e}$. $\mathcal{E}_{tr}$ and $\mathcal{E}_{ts}$ denote the set of training and test domains. 
In OOD prediction, the following holds: $|\mathcal{E}_{tr}|>1$ and $\mathcal{E}_{tr}\cap \mathcal{E}_{ts} = \emptyset$.
A domain is identified by its joint distribution $P(X^e,Y^e)$.
In this work, we do not restrict the causal model which generates the data.
We only require the joint distribution of a domain is obtained by a \emph{valid intervention} on the causal model.
An intervention is \emph{valid} iff it does not intervene on the label $Y$~\citep{ahuja2020empirical}.

\noindent\textbf{Problem Statement.}
%
%
Given data from multiple training domains $\{D_e\}_{e\in\mathcal{E}_{tr}}$, the goal is to predict the label $y_i^{e'}$ of each instance $i$ from a test domain $\{x_i^{e'},y_i^{e'}\}_{i=1}^{n_{e'}},e'\in\mathcal{E}_{ts}$.
Formally, we can define the problem of OOD prediction as an optimization problem~\citep{ye2021out}:
\begin{equation}
    \underset{\bm{\theta}_F,\bm{\theta}_G}{\arg\min} \underset{e\in\mathcal{E}_{ts}}{\sup} \mathds{E}_{(x,y)\sim D_e}[R^e(G(F(x)),y)].
    \label{eq:def_ood}
\end{equation}
In practice, previous work often uses a test domain which has significantly different distribution from the training to compute the supremum in Eq.~\eqref{eq:def_ood}~\citep{arjovsky2019invariant,krueger2020out}.
Note that without access to the test domain during the training phase in OOD prediction, we cannot compute the objective in Eq.~\eqref{eq:def_ood}. Instead, we have to propose objectives that can be computed with training data.

%


\noindent\textbf{Invariant Risk Minimization.} IRM~\citep{arjovsky2019invariant} is a recently proposed method to impose the inductive bias: the correlation between the invariant feature representations and the label should be robust across domains.
%
%
%
\cite{arjovsky2019invariant} present the original formulation of IRM as a two-stage optimization problem:
\begin{equation}
\begin{split} \underset{\bm{\theta}_F,\bm{\theta}_G}{\arg\min} & \sum_{e\in\mathcal{E}_{tr}}\mathds{E}_{(x,y)\sim D_e}[ R^e(G(F(x)),y)] \\
s.t. \;\; & \bm{\theta}_G \in \underset{\bm{\theta}'_G}{\arg\min} \;\; R^e(G(F(x);\bm{\theta}'_G),y), \forall e\in\mathcal{E}_{tr},
\\
\end{split}
\label{eq:irm}
\end{equation}
where $R^e$ denotes the loss function of domain $e$ (e.g., cross entropy loss).
Then, with the assumption that $G$ is a linear function and $R^e$ is convex, \cite{arjovsky2019invariant} show that the constraint in the optimization problem of IRM (Eq.~\eqref{eq:irm}) can be approximately imposed by adding a regularizer to the empirical risk as:
\begin{equation}
    \underset{\bm{\theta}_F}{\arg\min}\; \mathcal{L}_{IRM},
    \label{eq:IRM_v1_opt}
\end{equation}
where $\mathcal{L}_{IRM} = \sum_{e\in\mathcal{E}_{tr}} \mathcal{L}_{IRM}^e$ and
%
\begin{equation}
\begin{split}
    \mathcal{L}_{IRM}^e=\frac{1}{n_e}\sum_{i=1}^{n_e} R^e(F(x_i^e),y_i^e) \\ + \alpha||\bigtriangledown_{w|w=1.0}R^e(wF(x_i^e),y_i^e)||^2,
\end{split}
    \label{eq:IRM_v1}
\end{equation}
where $w=1.0$ is a scalar or a fixed dummy classifier, which replaces the linear predictor $G$ as shown in Theorem 4 of~\citep{arjovsky2019invariant}.
To distinguish from the two-stage form (Eq.~\eqref{eq:irm}), we will refer to the practical form of IRM (Eq.~\eqref{eq:IRM_v1_opt}) as the regularization form of IRM in the rest of this work.
%
%

\section{Desired and Undesired Solutions of IRM}

In this section, we show that the IRM can fail in the setting of strong triangle spuriousness as its solutions can come with non-overlapping spurious representations.
To empirically show it, we design a new dataset -- the colored MNIST plus (CMNIST+) to expose this limitation of IRM.
Then, we try to answer two crucial research questions: (1) why does IRM fail in CMNIST+? (2) Why does IRM work for the original colored MNIST (CMNIST) dataset?

%

%

\subsection{Why does IRM fail under strong triangle spuriousness?}
\label{subsec:irm_flaw}

\begin{figure}[tbh!]
 \centering
 \centering
 {\includegraphics[width=0.45\textwidth]{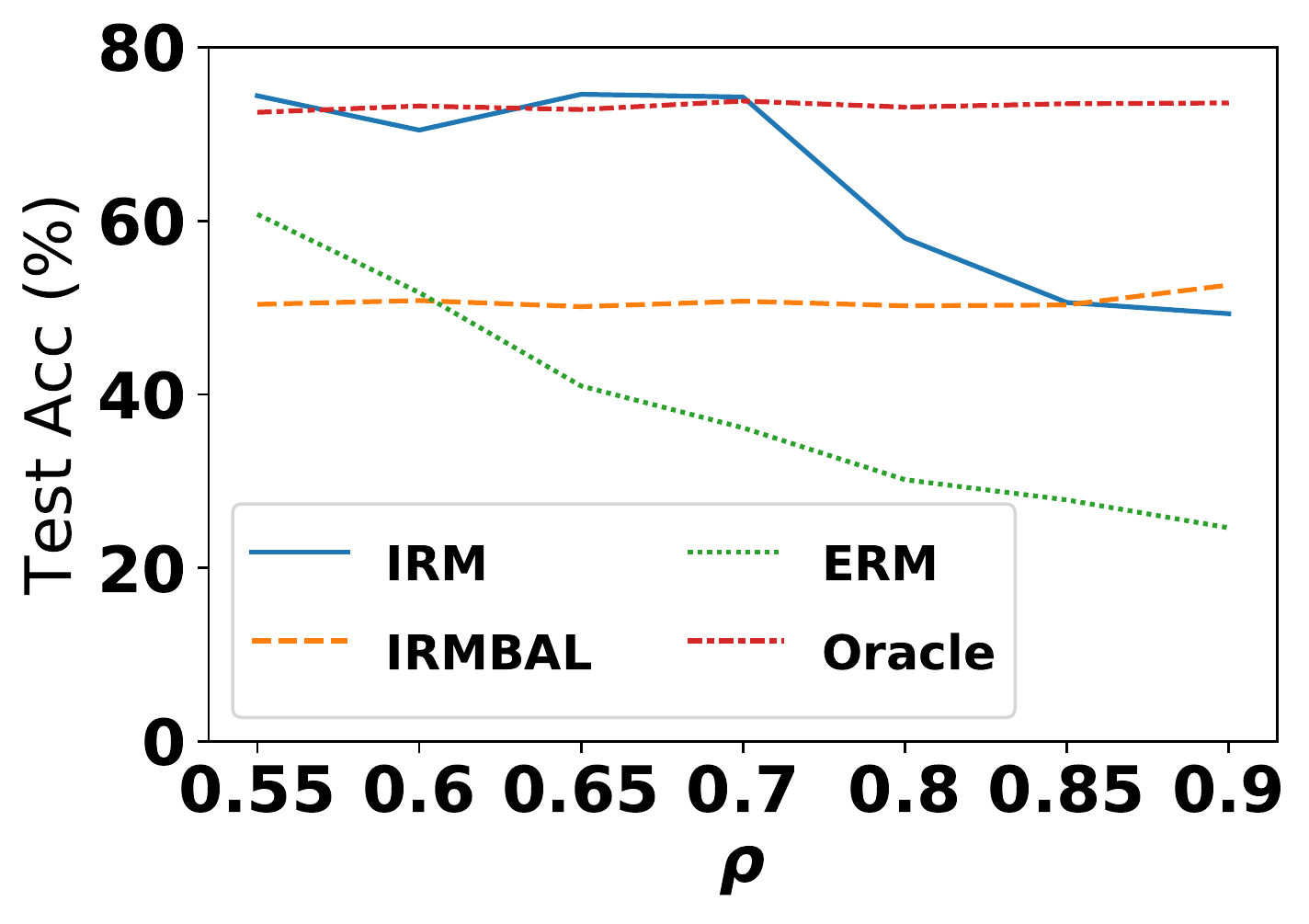}}
  \caption{Test domain accuracy of IRM,
  IRM with balanced classes in each domain (IRMBAL),
  ERM, and Oracle on CMNIST+. We can observe that, when the triangle spuriousness is strong ($\rho> 0.75$), the performance of IRM degrades dramatically. This is because, in this situation, DNNs get around the regularization of IRM with non-overlapping spurious feature representations across domains.
  The na\"ive solution, balancing classes in each domain cannot mitigate this problem of IRM.
  %
  } \label{fig:irm_cmnist_plus}
\end{figure}
Despite its empirical success in the CMNIST dataset~\citep{arjovsky2019invariant}, results in Fig.~\ref{fig:irm_cmnist_plus} show that IRM fails in our newly designed dataset -- colored MNIST plus (CMNIST+), under strong triangle spuriousness ($\rho> 0.75$).
%

%
%
%
%
%

\noindent\textbf{Accurately predicting the class label leads to desired solutions of IRM.}
%
First, we consider a desired solution that can lead to low values of the loss in Eq.~\eqref{eq:IRM_v1}. 
Specifically, we consider the case where there exists a representation that perfectly predicts the class label, i.e., $F(X)=Y$.
In this case, the representation $F(X)=Y$ results in low values of the regularizer in Eq.~\eqref{eq:IRM_v1}.
Intuitively, this can be justified by the fact that the correlation between $Y$ and itself must be invariant across different domains.
Formally, we show it with two widely used loss functions: binary cross entropy (BCE) and mean squared error (MSE).
For BCE, with $\mathcal{Y} = \{0,1\}$, we have:
\begin{equation}
\begin{split} & \bigtriangledown_{w|w=1.0,F(x_i^e)=y_i^e}R_{BCE}^e(wF(x_i^e),y_i^e) \\ & = - [y_i^e\frac{F(x_i^e)}{wF(x_i^e)} + (1-y_i^e)\frac{-F(x_i^e)}{1-wF(x_i^e)}] =0.
\end{split}
\label{eq:BCE_reg}
\end{equation}
For MSE, we have:
\begin{equation}
\begin{split} & \bigtriangledown_{w|w=1.0,F(x_i^e)=y_i^e}R_{MSE}^e(wF(x_i^e),y_i^e) \\ & = 2(wF(x_i^e)-y_i^e)F(x_i^e) = 0.
\end{split}
\label{eq:MSE_reg}
\end{equation}
These results also imply that with $F(X)\approx Y$, we can achieve low values of the IRM regularizer.
Note that $F(X)\approx Y$ also leads to low values of the empirical risk term in Eq.~\eqref{eq:IRM_v1}, which makes it a valid solution to the optimization problem of the regularization form of IRM (Eq.~\eqref{eq:IRM_v1_opt}).

\noindent\textbf{Undesired solutions of IRM pick up spurious correlations.}
However, as the functional relationship between $X$ and $Y$ is complicated in a vast majority of machine learning tasks, it is not likely to learn $F(X)\approx Y$.
Instead, $F(X)$ would extract features representations that are highly correlated with $Y$. 
Here, for simplicity, we consider three types of such features that can be picked up by $F(X)$: the invariant feature $S$ (e.g., shape), the spurious feature $C$ (e.g., color), and the domain variable $E$.
Then, under strong triangle spuriousness, we show that with spurious representations we can achieve low values of the IRM regularized loss.
%
%
%
%
In the setting of strong triangle spuriousness, when DNNs learn any non-overlapping representations $F(X)$, they circumvent the equality $\mathds{E}[Y|F(X),E=e] = \mathds{E}[Y|F(X),E=e'],e\not = e'$ imposed by the constraint in the original form of IRM (Eq.~\eqref{eq:irm}). This implies that $\mathds{E}[Y|F(X),E=e] \approx \mathds{E}[Y|F(X),E=e'],e\not = e'$ is imposed by the IRM regularized loss as a soft constraint.
This is because, with non-overlapping $F(X)$, for any data point in the representation space, $(F(x_i^e)=h,y_i^e)$, there does not exist $(F(x_j^{e'})=h,y_j^{e'})$, leaving the soft constraint ineffective.
However, this is not a desired behavior of IRM as non-overlapping representations $F(X)$ can pick up spurious features.
Here, we consider the non-overlapping representation $F(X)=E$.
Out of the three types of features ($F(X)=S$, $F(X)=C$, $F(X)=E$), only $F(X)=E$ would make $P(F(X)|E=e)$ and $P(F(X)|E=e'),e\not=e'$ completely non-overlapping.
%
%
%
To further justify that $F(X)=E$ is a good solution for the regularized form of IRM (Eq.~\eqref{eq:IRM_v1_opt}) under strong triangle spuriousness,
we find that $F(X)=E$ leads to low values of the empirical risk since the domain variable $E$ and the class label $Y$ hold a strong spurious correlation in this setting.
We can also justify this from a conditional independence perspective.
We know that $Y\indep E|F(X)$ is a necessary condition for any solution $F(X)$ of IRM (see Appendix~\ref{app:IRM_and_CI}).
Out of the three types of features, only $F(X) = E$ or $F(X) = S$ satisfies $Y\indep E|F(X)$. With strong triangle spuriousness, the model would learn $F(X)=E$ because the correlation between $E$ and $Y$ is stronger than that between $S$ and $Y$ in the training data.
%
%
Then, we describe the CMNIST+ dataset and show experimental and theoretical results to support our claim.

\noindent\textbf{The Colored MNIST Plus (CMNIST+) Dataset.}
We follow CMNIST~\citep{arjovsky2019invariant} to create CMNIST+ by resampling and adding colors to instances of MNIST.
Fig.~\ref{fig:causal_graph_cmnist_plus} shows the causal graphs that illustrate the data generating process of CMNIST+.
The relationships among $Y$, $C$ and $E$ are guaranteed to be spurious because a valid intervention on the variable $E$ would lead to changes in $P(Y|E)$ and $P(C|Y,E)$ without influencing the invariant correlation $P(Y|S)$.
The digits $0-4$ ($5-9$) are class $Y=1$ ($Y=0$). 
We randomly flip $25\%$ of class labels so that spuriousness can be stronger than the invariant correlation between the shape $S$ (invariant features) and the class label $Y$.
Table~\ref{tab:colored_mnist_plus} describes the dataset.
The variable $C\in\{G,B,R\}$ denotes the color, which represents the spurious feature. We explain why we use three colors for CMNIST+ in Appendix~\ref{app:data}. 
The parameter $\rho \in (0.5,1) $ controls the strength of the spurious correlations between the color (spurious features) $C$ and the class label $Y$ through the domain label $E$. The larger the value of $\rho$, the stronger the spurious correlations.
Intuitively, in the two training domains of CMNIST+, in addition to the strong spurious correlation between the class label $Y$ and color $C$, we set the spurious correlation between the domain label $E$ and color $C$ to be strong, too.
Thus, the CMNIST+ dataset can expose the problem of IRM: it would not penalize the DNN models that pick up the domain variable as the feature representation ($F(X) \approx E$).
To verify this claim, we show analysis results to support the experimental results in Fig.~\ref{fig:irm_cmnist_plus}.
The data generating process of CMNIST+ can be found in Append~\ref{app:data}.

\begin{table*}[tbh!]
\caption{Definition of CMNIST+}
\label{tab:colored_mnist_plus}
\centering
\small
\begin{tabular}{|c|c|c|c|c|c|}
\hline
$E$                      & $P(Y=1|E)$             & $Y$   & $P(C=G|Y,E)$ & $P(C=B|Y,E)$ & $P(C=R|Y,E)$ \\ \hline \hline
\multirow{2}{*}{$E=1$} & \multirow{2}{*}{0.9} & $Y=1$ & $\rho$          &  $(1-\rho)/2$        & $(1-\rho)/2$        \\ \cline{3-6} 
                       &                      & $Y=0$ & $(1-\rho)/2$         & $(1-\rho)/2$       & $\rho$        \\ \hline
\multirow{2}{*}{$E=2$} & \multirow{2}{*}{0.1} & $Y=1$ & $(1-\rho)/2$        & $\rho$         & $(1-\rho)/2$       \\ \cline{3-6} 
                       &                      & $Y=0$ & $(1-\rho)/2$         & $(1-\rho)/2$        & $\rho$        \\ \hline
\multirow{2}{*}{$E=3$} & \multirow{2}{*}{0.5} & $Y=1$ & 0.1          & 0.1         & 0.8        \\ \cline{3-6} 
                       &                      & $Y=0$ & 0.4          & 0.4         & 0.2        \\ \hline
\end{tabular}
\end{table*}

\begin{figure*}[tbh!]
\centering
\includegraphics[width=\textwidth]{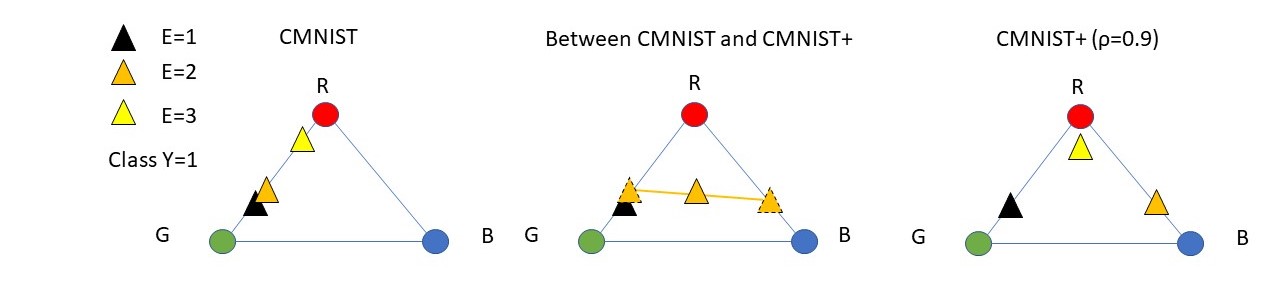}
\caption{We visualize the differences between CMNIST and CMNIST+ ($\rho=0.9$) in terms of $P(C|Y=1,E)$. Each large triangle represents the space of $P(C|Y,E)$. Each small triangle shows the values of $P(C=c|Y=1,E=e), c\in\{R,G,B\}, e\in\{1,2,3\}$. } \label{fig:cminst_plus_to_cmnist}
\end{figure*}

\noindent\textbf{Analytical Results for CMNIST+.}
Here, we show the analytical results for CMNIST+ to answer two questions: what features will be learned by ERM and IRM under under different triangle spuriousness? What is the expected test accuracy of these models?
%
%
For simplicity, we assume that the classifier is deterministic, which always predicts the majority class given the feature. If the number of instances from each class is the same, then it would predict a random label.
%
%
We analyze three types of spurious feature representations: (1) those that fit color, (2) those that fit the domain label and (3) those that use a combination of them for prediction.
The first case mimics the behavior of ERM that picks up the spurious correlation between color and the class label.
The second one represents a model satisfying IRM by $F(X)=E$.
The third one stands for a model satisfying IRM by $F(X)=Concat(E,C)$.
One may argue that IRM can be fixed if we simply balance the two classes in each domain.
Theoretical analysis shows that this is an invalid solution.
We summarize results in Table~\ref{tab:theo_res} where $\hat{E}$ is the domain label predicted by a deterministic classifier using color as the feature.
This is because in an unseen test domain, we cannot directly use the domain label for prediction.
The highlighted numbers show which type of features would be learned by ERM and IRM.
As $\rho$ increases, ERM and IRM are more likely to fit spurious features.
This explains results in Fig.~\ref{fig:irm_cmnist_plus}.
As a DNN may still pick up some invariant features (shape in CMNIST+) even when $\rho=0.9$, the test accuracy of IRM in practice would be greater than $0.35$ without label balancing and $0.2$ with label balancing.
Similarly, the test accuracy of ERM would be greater than $0.2$.
From Fig.~\ref{fig:irm_cmnist_plus}, we can see the test accuracy of IRM, IRM with label balancing and ERM are slightly better than the aforementioned lower bounds.
The derivations can be found in Appendix~\ref{app:theo}.

\noindent\textbf{Experimental Setup.}
%
We use a slightly modified LeNet-5 ~\citep{lecun1998gradient} instead of a three-layer MLP because LeNet-5 has more predictive power such that it can pick up the three types of spurious correlations (color -- domain, color -- class, and domain -- class) or the invariant relationship (shape -- class).
To take input with three colors, we set the first CNN layer to have three channels. 
We randomly split the instances from the training domains into $80\%$ training and $20\%$ validation.
The model selection is done by picking the one with the lowest validation loss in each run~\citep{gulrajani2020search}.
We report the average test accuracy of the selected models in 10 runs.
It is crucial to ensure only data from training domains are used in model selection. This is because, in OOD prediction, the test domain is assumed to be unknown during training and validation~\citep{gulrajani2020search}.
%
During training, we begin applying the IRM penalty at iteration $K_{IRM} \in \{200, 400, 600, 800, 1000\}$.
%
%
By varying $K_{IRM}$, we aim to examine the following hypothesis: IRM works by pushing the spurious features out of the representations learned by the standard ERM training before the IRM penalty is applied~\citep{krueger2020out}.
More details on the setup can be found in Appendix~\ref{app:setup}. 

\noindent\textbf{Experimental Results.} Fig.~\ref{fig:irm_cmnist_plus} shows the performance of IRM and IRM with balanced classes in each domain (IRMBAL). We make the following observations: first, the test accuracy of IRM drops dramatically under strong triangle spuriousness.
Second, we show that a na\"ive fix for IRM, balancing the two classes in each domain by oversampling the minority class, does not improve the performance.

\begin{table*}[tbh!]
\caption{Analytical results. Boldface numbers show (1) what type of feature would be learned by ERM and IRM and (2) what the corresponding test accuracy would be for CMNIST+. We show the validation/test accuracy with the three types of spurious feature representations (the color C, the domain E and the concatenation of E and C) and those with the invariant feature representation (S: shape) on CMNIST+. The ones with the best validation accuracy in each setting are highlighted, which imply the type of representations would be learned by ERM and IRM.}
\label{tab:theo_res}
\small
\centering
\begin{tabular}{|c|c|c|c|c|c|c|c|c|}
\hline
\multicolumn{9}{|c|}{Validation/Test Accuracy}                                                                                                                                             \\ \hline
       & \multicolumn{4}{c|}{without label balancing}                                            & \multicolumn{4}{c|}{with label balancing}                                               \\ \hline
$\rho$ & $\hat{P}(Y|C)$     & $\hat{P}(Y|\hat{E})$ & $\hat{P}(Y|\hat{E},C)$ & $\hat{P}(Y|S)$   & $\hat{P}(Y|C)$     & $\hat{P}(Y|\hat{E})$ & $\hat{P}(Y|\hat{E},C)$ & $\hat{P}(Y|S)$   \\ \hline
0.55   & 0.662/0.2          & 0.646/0.35           & 0.662/0.35             & \textbf{0.75/0.75} & 0.662/0.2          & 0.5/0.5              & 0.662/0.2              & \textbf{0.75/0.75} \\ \hline
0.6    & 0.7/0.2            & 0.68/0.35            & 0.7/0.35               & \textbf{0.75/0.75} & 0.7/0.2            & 0.5/0.5              & 0.7/0.2                & \textbf{0.75/0.75} \\ \hline
0.65   & 0.738/0.2          & 0.714/0.35           & 0.738/0.35             & \textbf{0.75/0.75} & 0.738/0.2          & 0.5/0.5              & 0.737/0.2              & \textbf{0.75/0.75} \\ \hline
0.7    & \textbf{0.775/0.2} & 0.748/0.35           & \textbf{0.775/0.35}    & 0.75/0.75          & \textbf{0.775/0.2} & 0.5/0.5              & \textbf{0.775/0.2}     & 0.75/0.75          \\ \hline
0.8    & \textbf{0.85/0.2}  & 0.815/0.35           & 0.815/0.35     & 0.75/0.75          & \textbf{0.85/0.2}  & 0.5/0.5              & \textbf{0.85/0.2}      & 0.75/0.75          \\ \hline
0.85   & \textbf{0.888/0.2} & 0.849/0.35           & 0.849/0.2              & 0.75/0.75          & \textbf{0.888/0.2} & 0.5/0.5              & \textbf{0.888/0.2}     & 0.75/0.75          \\ \hline
0.9    & \textbf{0.925/0.2} & 0.883/0.35           & 0.883/0.2              & 0.75/0.75          & \textbf{0.925/0.2} & 0.5/0.5              & \textbf{0.925/0.2}     & 0.75/0.75          \\ \hline
\end{tabular}
\end{table*}

\subsection{Why does IRM work for the Original Colored MNIST?}
%

\begin{figure}[tbh!]
 \centering
 \begin{minipage}{0.32\textwidth}
 \centering
  \subfigure[IRM on CMNIST test set]
 {\includegraphics[width=\textwidth]{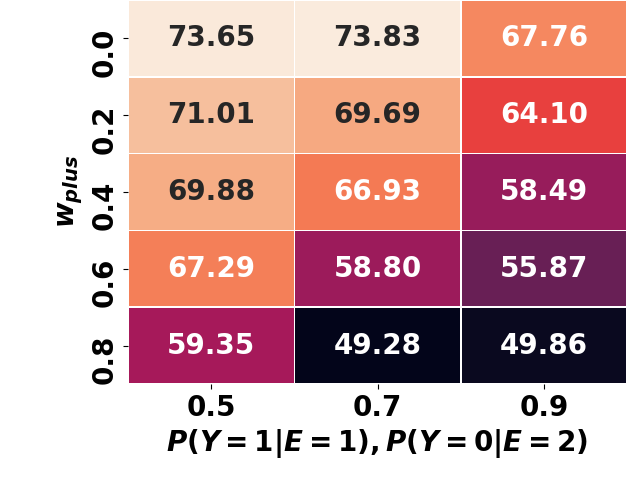}}
 \end{minipage}
  \hfil
  \begin{minipage}{0.32\textwidth}
  \centering
  \subfigure[IRM on CMNIST+ test set]
  {\label{fig:proposed}\includegraphics[width=\textwidth]{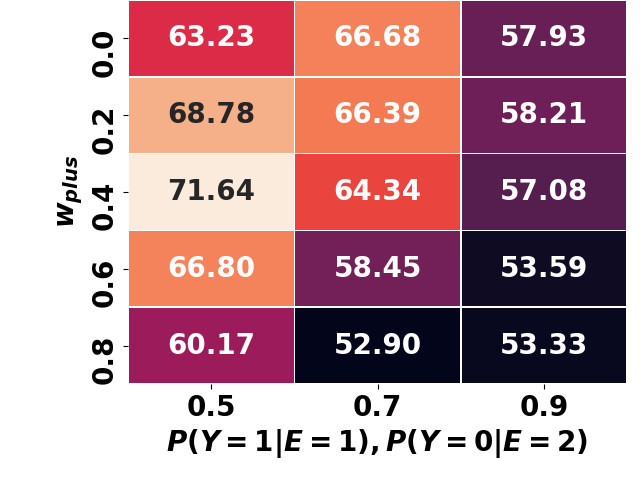}}
  \end{minipage}
  \caption{Accuracy of IRM trained on datasets between CMNIST and CMNIST+ and tested on the test sets of CMNIST and CMNIST+ ($w_{plus}$ increases): As the training set becomes more similar to CMNIST+, the accuracy of IRM gradually drops, on the test sets of CMNIST and CMNIST+.} \label{fig:irm_cmnist_trend}
\end{figure}

The Colored MNIST (CMNIST) dataset cannot expose the limitation of IRM under strong triangle spuriousness.
This is because its two training domains are quite similar.
As shown in the large triangle on the left in Fig.~\ref{fig:cminst_plus_to_cmnist}, the values of $P(C|Y,E)$ are similar for $E=1$ and $E=2$.
In addition, the values of $P(Y|E)$ are the same for all $E$.
This makes it difficult make the value of the IRM regularizer low by learning feature representations $F(X)\approx E$.
%
%

Then, we present experiments to show how IRM gradually goes from working on CMNIST to failing on CMNIST+.
In Fig.~\ref{fig:cminst_plus_to_cmnist} Left and Right, we observe the differences between between CMNIST and CMNIST+ ($\rho=0.9$).
%
In Fig.~\ref{fig:cminst_plus_to_cmnist} Middle, we create various datasets that interpolate between these two datasets, illustrated by the yellow line in the middle triangle. 
We use a parameter $w_{plus}\in[0,1]$ to control $P(C|Y,E)$ of the weights of  CMNIST+ ($\rho=0.9$) in the interpolated dataset:
\begin{equation}
\begin{split}
       & P(C|Y,E) = \\ & \frac{P_{\text{cmnist+}}(C|Y,E)w_{plus} + P_{\text{cmnist}}(C|Y,E)(1-w_{plus})}{\sum_{C} (P_{\text{cmnist+}}(C|Y,E)w_{plus}+P_{\text{cmnist}}(C|Y,E)(1-w_{plus}))},
\end{split}
\end{equation}
where $P_{\text{cmnist+}}(C|Y,E)$ and $P_{\text{cmnist}}(C|Y,E)$ are the values of $P(C|Y,E)$ in CMNIST+ and CMNIST.
When $w_{plus}=0$ and $P(Y=1|E=1)=P(Y=0|E=2)=0.5$, the dataset is the same with CMNIST.
As $w_{plus}$ and $P(Y=1|E=1)=P(Y=0|E=2)$ increase, the dataset becomes more similar to CMNIST+, and becomes the same with CMNIST+ ($\rho=0.9$) when $w_{plus} = 1$ and $P(Y=1|E=1)=P(Y=0|E=2) = 0.9$. 
For the training sets, we set $P(Y=1|E=1) = P(Y=0|E=2) \in \{0.5,0.7,0.9\}$.
Fig.~\ref{fig:irm_cmnist_trend} shows results of IRM on the test sets of both CMNIST and CMNIST+.
The performance of IRM gradually drops when the values of $P(C|Y,E)$ becomes more similar to CMNIST+ (when the triangle spuriousness becomes stronger).

\section{An Effective Fix for IRM}
In this section, we propose a simple but effective solution to address the aforementioned non-overlapping issue of IRM.
In a series of experiments, we show that the proposed method can improve the performance of IRM even under strong triangle spuriousness.

Since the IRM regularization can lead to undesired solutions, we propose an effective solution -- combining the conditional distribution matching (CDM) regularization and the IRM regularization.
The CDM regularization~\citep{li2018deep,long2018conditional} aims to impose the soft constraint, $P(F(X)|Y,E)\approx P(F(X)|Y)$. In other words, the CDM regularizer pushes feature representation distribution of instances from the same class to be similar across different domains.
We first explain why this can be a reasonable solution.
Then, we propose two types of models that combine the two constraints.

%


%
%
Here, we first explain why combining the regularization of IRM with that of CDM can be an effective fix for the former. 

\textbf{Accurately predicting the label satisfies CDM.} In this case, we consider the representation $F(X)\approx Y$. From Section~\ref{subsec:irm_flaw}, we know that the IRM regularization takes low values with $F(X)\approx Y$.
The CDM condition also approximately holds as $P(F(X)|Y,E) \approx P(Y|Y,E) = P(Y|Y)\approx P(F(X)|Y)$.

\textbf{Accurately predicting the domain label would be penalized by CDM.}
In this case, $F(X)\approx E$ and the IRM regularization can be circumvented. However, the CDM soft constraint does not hold as $P(F(X)|Y,E)\approx P(E|Y,E) \not \approx P(E|Y) \approx P(F(X)|Y))$. This case implies that we can add CDM to IRM to exclude the undesirable solution of learning spurious features that accurately predict the domain label.

One way to enforce the CDM regularization is through adversarial training: a discriminator tries to infer the source domain from the feature representation $F(X)$ and $F(X)$ tries to adjust itself to fool the discriminator. From this perspective, we can find that the CDM regularization literally prevents $F(X)$ from making use of the domain labels to achieve high training accuracy (i.e., low $R_e$). 
To impose the CDM regularization, we aim to minimize the divergence between conditional representation distributions from different domains. The divergence can be denoted as $div(P(F(X)|Y,E=e)||P(F(X)|Y,E=e')),e\not=e'$.
%

We propose to use two implementations of this divergence -- Maximum Mean Discrepancy (MMD) and Kullback-Leibler (KL) divergence (through adversarial training). Thus, we end up with two algorithms: IRM-MMD and IRM-ACDM (i.e., IRM-Adversarial Conditional Distribution Matching).

\noindent\textbf{IRM-MMD.}
%
%
In IRM-MMD, we adopt Maximum Mean Discrepancy (MMD)~\citep{long2015learning,tolstikhin2017wasserstein,shalit2017estimating} as the distribution divergence.
The MMD between  $P$ and $Q$, two $d$-dimensional distributions of feature representations, can be defined as:
\begin{equation}
    MMD_k(P,Q) = \underset{f\in\mathcal{H}}{\sup}|\mathds{E}_{Z\sim P}[f(Z)] - \mathds{E}_{Z\sim Q}[f(Z)]|,
    \label{eq:MMD}
\end{equation}
where $Z\in \mathds{R}^d$, $f:\mathds{R}^d\rightarrow\mathds{R}$ maps a feature representation to a real value, $k:\mathds{R}^d\times\mathds{R}^d \rightarrow \mathds{R}$ denotes the characteristic kernel of $f$ and $\mathcal{H}$ is the RHKS of $k$.
The MMD in Eq.~\ref{eq:MMD} is not directly computable.
So, we use the unbiased estimator of MMD~\citep{gretton2012kernel} with $N$ samples $z_1^P,...,z_N^P$ from $P$ and $M$ samples $z_1^Q,...,z_M^Q$ from $Q$:
\begin{equation}
\begin{split}
        & MMD_k(P,Q) = \frac{1}{N(N-1)}\sum_{i\not=j}k(z_i^P,z_j^P) +\\ & \frac{1}{M(M-1)}\sum_{i\not=j}k(z_i^Q,z_j^Q) - \frac{2}{MN}\sum_{i=1}^N\sum_{j=1}^M k(z_i^P,z_j^Q).
\end{split}
\end{equation}
With MMD defined, we can define the loss function of IRM-MMD as:
\begin{equation}
\begin{split}
 & \underset{\bm{\theta}}{\arg\min} \sum_{e\in\mathcal{E}_{tr}} \mathcal{L}^e_{IRM} + \\ & \beta\sum_{y\in\mathcal{Y}}\sum_{e\in\mathcal{E}_{tr}}\sum_{e'\in\mathcal{E}_{tr}\setminus e} MMD_k(P(F(X)|y,e),P(F(X)|y,e')),
\end{split}
\end{equation}
where $\mathcal{L}_{IRM}^e$ is the IRM regularized loss~\citep{arjovsky2019invariant} for domain $e$ (see Eq.~\ref{eq:IRM_v1}), hyperparameters $\alpha$ and $\beta$  control the trade-off between the main loss, the IRM constraint and the CDM constraint.

\noindent\textbf{IRM-ACDM.}
In IRM-ACDM, we make $div(P(F(X)|Y,E=e)||P(F(X)|Y,E=e'))$ be the Kullback-Leibler (KL) divergence and use adversarial learning to estimate it~\citep{tolstikhin2017wasserstein,song2020novel}.
More precisely, we define the loss function of IRM-ACDM as:
\begin{equation}
\begin{split}
    & \underset{\bm{\theta}}{\arg\min} \sum_{e\in\mathcal{E}_{tr}} \mathcal{L}^e_{IRM} + \\ & \beta\sum_{y\in\mathcal{Y}} \sum_{e\in\mathcal{E}_{tr}} \gamma_e^y KL(P(F(X)|Y=y,E=e)||P(F(X)|Y=y)),
\end{split}
\end{equation}
where $\gamma_e^y := P(E=e, Y=y)$. In adversarial learning, a conditional discriminator $D:\mathds{R}^d\times\mathcal{Y}\rightarrow \mathcal{E}_{tr}$ with parameters $\bm{\theta}_D$ is introduced to predict the domain label of an instance, given its feature representation and class label.
%
As proved in \citep{li2018deep,song2020novel}, the KL divergence above can be estimated through the minimax game below:
\begin{equation}
\begin{split}
 & \underset{\bm{\theta}}{\min}    \;\underset{\bm{\theta}_D}{\max} \sum_{e\in\mathcal{E}_{tr}} \mathcal{L}^e_{IRM} + \\ & \beta\sum_{y\in\mathcal{Y}} \sum_{e\in\mathcal{E}_{tr}} \gamma_e^y \mathds{E}_{F(x)\sim P(F(X)|Y=y,E=e)} [\log D^e(F(x),y)],
\end{split}
    \label{eq:KL}
\end{equation}
where $D^e(F(x),y) = \hat{P}(E=e|F(x),y)$ is the predicted probability of the instance for domain $e$ by the discriminator $D$. In practice, we solve the minimax game above efficiently by the alternative gradient ascent/descent algorithm. 
%
%

\noindent\textbf{Experimental Results.} We evaluate IRM-MMD and IRM-ACDM on CMNIST+ to show their efficacy under strong triangle spuriousness. 
We let the output of the second last layer of LeNet-5 to be the feature representation $F(X)$.
For IRM-MMD, we use the multiple kernel MMD (MKMMD)~\citep{gretton2012optimal,long2015learning}. For IRM-ACDM, we use a two-layer MLP with $|\mathcal{E}_{tr}|$ outputs as the discriminator $D$.
To show that CDM alone cannot solve OOD prediction under strong triangle spuriousness,
we set the weight of the IRM regularizer $\alpha=0$ in IRM-MMD and IRM-ACDM to obtain the ERM models regularized by MMD and ACDM, respectively.
In addition, we consider EIIL~\citep{creager2020environment} that drops the observed domain labels and learn soft domain labels for each instance by solving a minimax game. In the max step, it learns soft domain labels for instances s.t. the IRM regularized loss is maximized.
In the min step, it minimizes the IRM regularized loss. 
We also include ERM and oracle as our baselines.
The oracle uses the original LeNet-5 architecture with a single-channel CNN as the first layer.
It is trained and tested with instances transformed into grayscale.

Table~\ref{tab:irm_mmd_acdm_cmnist_plus} shows the performance of IRM-MMD, IRM-ACDM and the baselines.
Under strong triangle spuriousness, compared to IRM (Fig.~\ref{fig:irm_cmnist_plus}), we can observe the significant and consistent performance improvement over IRM resulting from combining CDM with IRM regularization.
IRM-MMD and IRM-ACDM also outperform MMD and ACDM. This verifies that CDM alone cannot solve the OOD prediction problem.
EIIL can only achieve a worse performance than IRM.
This implies that the information of the domain variable plays a crucial role in learning invariant features.
Compared to learning the domain variable as parameter (EIIL), models can better distinguish invariant features from spurious ones with the information of the domain variable on which valid interventions generate different domains.
%



\begin{table}[h]
\caption{Test domain accuracy on CMNIST+: IRM-CDM significantly and consistently outperforms the baselines and ablation models with $\rho \in \{0.8,0.85\}$.} \label{tab:irm_mmd_acdm_cmnist_plus}
\centering
\small
\begin{tabular}{|c|c|c|c|}
\hline
Method   & $\rho= 0.8$              & $\rho=0.85  $           & $\rho=0.9  $            \\ \hline
IRM-MMD (ours)  & \textbf{67.08\%} & \textbf{58.47\%} & 49.02\% \\ 
IRM-ACDM (ours) & \textbf{70.15\%} & \textbf{63.91\%} & 49.28\% \\
 \hline IRM      & 58.00\%          & 50.58\%          & 49.29\%          \\
\hline EIIL      & 43.40\%          & 43.24\%          & 40.93\%          \\
\hline
MMD      & 23.04\%          & 25.22\%          & 24.22\%          \\ 
ACDM     & 30.41\%          & 29.48\%          & 25.53\%          \\ \hline
ERM      & 30.16\%          & 27.83\%          & 24.61\%          \\ \hline
Oracle   & 73.10\%          & 73.49\%          & 73.58\%          \\ \hline
\end{tabular}
\end{table}

\section{Concluding Remarks}

This work focuses on addressing the issue of IRM for the OOD prediction problem under strong triangle spuriousness.
Strong triangle spuriousness means the correlations between spurious features, the domain variable and the class label are stronger than the invariant correlation between invariant features and the label.
We find an important limitation of IRM in OOD prediction under strong triangle spuriousness: it can be circumvented by spurious feature representations that are predictive of the domain label.
To verify it, we design the CMNIST+ dataset which has strong triangle spuriousness between color (spurious features) the class label through their common cause -- the domain variable.
On CMNIST+, we observe the performance of IRM dramatically drops when the triangle spuriousness becomes stronger.
Based on this observation, we propose a simple but an effective fix to mitigate this issue of IRM.
The proposed approach combines CDM and IRM because CDM can also be satisfied by causal feature representations. At the same time, CDM can prevent DNNs from picking up the aforementioned spurious feature representations.
Experimental results on CMNIST+ show significant performance improvement of the proposed method, demonstrating its effectiveness.
Interesting future work includes (1) extension of the proposed method to OOD prediction tasks in complex data (e.g., graphs and time series) and (2) development of general causal inductive bias that can impose various conditional independence.

\clearpage

\bibliographystyle{plain}
\bibliography{uai2021-template}

\clearpage
\appendix

\section{IRM and the Conditional Independence}
\label{app:IRM_and_CI}
Under general conditions, the conditional independence $Y \indep E |F(X)$ is a necessary condition for solutions of the original IRM optimization problem.
In~\citep{arjovsky2019invariant}, IRM is defined as a two-stage optimization problem.
Any solution $F(X)$ to the IRM optimization problem must satisfy $Y \indep E |F(X)$.
However, some $F(X)$ satisfying $Y \indep E |F(X)$ may not be a solution to the original IRM problem.
For example, generally, $F(X) = E$ would not minimize the sum of the domain-specific risk $R^e$.
However, under strong $\Lambda$ spuriousness, there exist solutions to the original IRM problem that still pick up spurious features.
Consider the extreme case, in the training data, if $Y=E$, then $F(X)=E$ is a solution to the problem.


\section{Analysis for CMNIST+}
\label{app:theo}
\noindent\textit{Fitting Color for Classification.}
In expectation, the ERM model would learn color as the feature representation for classification.
Here, we show the theoretical results for such cases.
When the model learns color $C$ as the feature representation $F(X)$, we have:
\begin{equation}
    \begin{split}
        P(Y|C) & = \sum_{E} P(Y|C,E)P(E|C) \\
         & = \sum_{E}  \frac{P(C|Y,E)P(Y|E)}{P(C|E)}P(E|C)
         \\ & = \sum_{E}  P(C|Y,E)P(Y|E)\frac{P(E)}{P(C)}
    \end{split}
    \label{eq:py_c}
\end{equation}
The second equality is by Bayes' rule.
We know that $P(C) = \sum_{Y}\sum_{E}P(C|Y,E)P(Y|E)P(E)$.
Then, given $P(C|Y,E)$, $P(Y|E)$, $P(C)$ and Eq.~\ref{eq:py_c}, 
we can obtain $P(Y|C)$ as shown in Table~\ref{tab:py_c}.
These results imply that the deterministic classifier $\hat{P}(Y=1|C=G)=1$, $\hat{P}(Y=1|C=B)=1$, $\hat{P}(Y=1|C=R)=0$ for all $\rho\in[0.55,0.9]$ as shown in Table~\ref{tab:py_c}.
So, the accuracy of the deterministic classifier $\hat{P}(Y|C)$ on the test set is $0.2$.
Its accuracy on the training set is $\sum_C P(\hat{Y}=Y|C)P(C)$ where $P(C)=\sum_{E}\sum_{Y}P(C|Y,E)P(Y|E)P(E)$.

\noindent\textit{Fitting Domain Label for Classification.} Since the spurious correlation between the domain label and the class label is strong, and IRM cannot penalize models fitting the domain label, $P(Y|E)$ can help us understand the expected behavior of the IRM model.
Note that the test data is from an unseen domain. So, we analyze the model that first predicts domain by color, then predicts class label by domain.
First, we analyze $P(E|C)$ as below:
\begin{equation}
    \begin{split}
        P(E|C) & = \sum_{Y}P(E|C,Y)P(Y|C) \\
        & = \sum_Y \frac{P(C|Y,E)P(E|Y)P(Y|C)}{P(C|Y)} \\ 
        & = \sum_Y \frac{P(C|Y,E)P(E|Y)P(Y)}{P(C)} \\
        &= \sum_Y \frac{P(C|Y,E)P(Y|E)P(E)}{P(C)},
    \end{split}
    \label{eq:pe_c}
\end{equation}
where the second and forth qualities are by Bayes' rule.
With Eq.~\ref{eq:pe_c}, we can list the values of $P(E|C)$ in Table~\ref{tab:pe_c} for $\rho\in[0.55,0.9]$.
Thus, we know the deterministic domain prediction results would be $\hat{P}(E=1|C=G)=1$, $\hat{P}(E=1|C=B)=0$, and $\hat{P}(E=1|C=R)=0$.
Since $P(Y|E)$ is given in Table~\ref{tab:colored_mnist_plus}, we can obtain the deterministic classifier's predictions as:
\begin{equation}
    \hat{P}(Y=1|C=G)=1, \; \hat{P}(Y=1|C=B)=0, \; \hat{P}(Y=1|C=R)=0.
    \label{eq:py_hat_e}
\end{equation}
So, the expected test accuracy of the model would be $0.35$.
Recall that the model first predicts domain label by color and then predict class label by the predicted domain.
By doing this, it would have $F(X)\approx E$. This makes it approximately satisfy the IRM constraint $Y\indep E |F(X)$ since $Y\indep E | E$. This implies that the model's performance can be treated as the expected performance of the IRM model in CMNIST+.
In terms of the performance of $\hat{P}(Y|\hat{E})$ on the training set, we can get the results using the same prediction rules as in Eq.~\ref{eq:py_hat_e}.

One may argue that the IRM model can perform well if we balance the two classes in each domain.
Here, we theoretically show this is not the case.
By setting $P(Y|E)=0.5$, we can obtain the values of $P(E|C)$ in Table~\ref{tab:pe_c_bal}.
Note that practically this can be done by oversampling the minority class of each domain in each mini-batch.
However, since $P(Y|E)=0.5$, the predictions made by the deterministic classifier $\hat{P}(Y|E)$ would be just random guess, leading to a test accuracy of $0.5$.

\noindent\textit{Fitting both Domain and Color for Classification.}
Here, we consider the model that first predicts the domain label by color and then predicts the class label by both the color and the predicted domain label.
The first step is the same as the model fitting the domain label.
For the second step, we analyze $P(Y|C,E)$ as below:
\begin{equation}
    \begin{split}
        P(Y|C,E) = \frac{P(C|Y,E)P(Y|E)}{P(C|E)}.
    \end{split}
    \label{eq:py_ce}
\end{equation}
With $P(C|E) = \sum_Y P(C|Y,E)P(Y|E)$ and Eq.~\ref{eq:py_ce}, we obtain values of $P(Y|C,E)$ as shown in Table~\ref{tab:py_ce}. 
So, given the predicted domains, $\hat{P}(E=1|C=G)=1$, $\hat{P}(E=1|C=B)=0$, and $\hat{P}(E=1|C=R)=0$, the predictions on the class label are $\hat{P}(Y=1|C=G,E=1)=1$, $\hat{P}(Y=1|C=B,E=2)=1, \; \rho > 0.8$, $\hat{P}(Y=1|C=B,E=2)=0, \; \rho \le 0.8$ and $\hat{P}(Y=1|C=R,E=2)=0$.
So, the test accuracy is $0.35$ when $\rho \le 0.8$ and $0.2$ when $\rho > 0.8$.

Similarly, when we make $P(Y|E)=0.5$ by oversampling the minority class in each domain, we can obtain the predictions shown in Table~\ref{tab:py_ce_bal}.
So, the deterministic model would make predictions as $\hat{P}(Y=1|C=G,E=1)=1$, $\hat{P}(Y=1|C=B,E=2)=1$, and $\hat{P}(Y=1|C=R,E)=0$.
This would lead to a test accuracy of $0.2$.
These results reflect the reasons why the IRM model trained with the balanced classes in each domain ($P(Y|E)=0.5$) has worse performance compared to its counterpart trained with the original CMNIST+ data.

\begin{table}[tbh!]
\caption{Analysis results: fitting color}
\label{tab:py_c}
\centering
\small
\begin{tabular}{|c|c|c|c|}
\hline
$\rho$ & $P(Y=1|C=G)$ & $P(Y=1|C=B)$ & $P(Y=1|C=R)$ \\ \hline
0.55        & 0.697        & 0.534       & 0.29       \\ \hline
0.6         & 0.737        & 0.545       & 0.25       \\ \hline
0.65        & 0.775        & 0.56        & 0.212      \\ \hline
0.7         & 0.811        & 0.577       & 0.176      \\ \hline
0.8         & 0.88         & 0.63        & 0.111      \\ \hline
0.85        & 0.912        & 0.67        & 0.081      \\ \hline
0.9         & 0.942        & 0.73        & 0.053      \\ \hline
\end{tabular}
\end{table}

\begin{table}[tbh!]
\caption{Analysis results: fitting color, when $P(Y|E)=0.5$.}
\label{tab:py_c_bal}
\small
\centering
\begin{tabular}{|c|c|c|c|}
\hline
$\rho$ & $P(Y=1|C=G)$ & $P(Y=1|C=B)$ & $P(Y=1|C=R)$ \\ \hline
0.55        & 0.633        & 0.633       & 0.29       \\ \hline
0.6         & 0.667        & 0.667       & 0.25       \\ \hline
0.65        & 0.702        & 0.702       & 0.212      \\ \hline
0.7         & 0.739        & 0.739       & 0.176      \\ \hline
0.8         & 0.818        & 0.818       & 0.111      \\ \hline
0.85        & 0.86         & 0.86        & 0.081      \\ \hline
0.9         & 0.905        & 0.905       & 0.053      \\ \hline
\end{tabular}
\end{table}

\begin{table}[tbh!]
\caption{Theoretical analysis results: predicting domain by color.}
\label{tab:pe_c}
\centering
\small
\begin{tabular}{|c|c|c|c|}
\hline
$\rho$ & $P(E=1|C=G)$ & $P(E=1|C=B)$ & $P(E=1|C=R)$ \\ \hline
0.55        & 0.697        & 0.466       & 0.332      \\ \hline
0.6         & 0.737        & 0.455       & 0.3        \\ \hline
0.65        & 0.775        & 0.44        & 0.27       \\ \hline
0.7         & 0.811        & 0.423       & 0.241      \\ \hline
0.8         & 0.88         & 0.37        & 0.189      \\ \hline
0.85        & 0.912        & 0.33        & 0.165      \\ \hline
0.9         & 0.942        & 0.27        & 0.142      \\ \hline
\end{tabular}
\end{table}

\begin{table}[tbh!]
\caption{Theoretical analysis results: predicting domain by color, when $P(Y|E)=0.5$.}
\label{tab:pe_c_bal}
\centering
\small
\begin{tabular}{|c|c|c|c|}
\hline
$\rho$ & $P(E=1|C=G)$ & $P(E=1|C=B)$ & $P(E=1|C=R)$ \\ \hline
0.55        & 0.633        & 0.367       & 0.5        \\ \hline
0.6         & 0.667        & 0.333       & 0.5        \\ \hline
0.65        & 0.702        & 0.298       & 0.5        \\ \hline
0.7         & 0.739        & 0.261       & 0.5        \\ \hline
0.8         & 0.818        & 0.182       & 0.5        \\ \hline
0.85        & 0.86         & 0.14        & 0.5        \\ \hline
0.9         & 0.905        & 0.095       & 0.5        \\ \hline
\end{tabular}
\end{table}

\begin{table*}[tbh!]
\small
\caption{Predicting by both color and domain}
\label{tab:py_ce}
\centering
\begin{tabular}{|c|c|c|c|c|c|c|}
\hline
$\rho$ & $P(Y=1|G,1)$ & $P(Y=1|G,2)$ & $P(Y=1|B,1)$ & $P(Y=1|B,2)$ & $P(Y=1|R,1)$ & $P(Y=1|R,2)$ \\ \hline
0.55        & 0.957             & 0.1               & 0.9              & 0.214            & 0.786           & 0.043           \\ \hline
0.6         & 0.964             & 0.1               & 0.9              & 0.25             & 0.75            & 0.036           \\ \hline
0.65        & 0.971             & 0.1               & 0.9              & 0.292            & 0.708           & 0.029           \\ \hline
0.7         & 0.977             & 0.1               & 0.9              & 0.341            & 0.659           & 0.023           \\ \hline
0.8         & 0.986             & 0.1               & 0.9              & 0.471            & 0.529           & 0.014           \\ \hline
0.85        & 0.99              & 0.1               & 0.9              & 0.557            & 0.443           & 0.01            \\ \hline
0.9         & 0.994             & 0.1               & 0.9              & 0.667            & 0.333           & 0.006           \\ \hline
\end{tabular}
\end{table*}

\begin{table*}[tbh!]
\centering
\small
\caption{Predicting by both color and domain, class balanced}
\label{tab:py_ce_bal}
\begin{tabular}{|c|c|c|c|c|c|c|}
\hline
$\rho$ & $P(Y=1|G,1)$ & $P(Y=1|G,2)$ & $P(Y=1|B,1)$ & $P(Y=1|B,2)$ & $P(Y=1|R,1)$ & $P(Y=1|R,2)$ \\ \hline
0.55        & 0.71              & 0.5               & 0.5              & 0.71             & 0.29            & 0.29            \\ \hline
0.6         & 0.75              & 0.5               & 0.5              & 0.75             & 0.25            & 0.25            \\ \hline
0.65        & 0.788             & 0.5               & 0.5              & 0.788            & 0.212           & 0.212           \\ \hline
0.7         & 0.824             & 0.5               & 0.5              & 0.824            & 0.176           & 0.176           \\ \hline
0.8         & 0.889             & 0.5               & 0.5              & 0.889            & 0.111           & 0.111           \\ \hline
0.85        & 0.919             & 0.5               & 0.5              & 0.919            & 0.081           & 0.081           \\ \hline
0.9         & 0.947             & 0.5               & 0.5              & 0.947            & 0.053           & 0.053           \\ \hline
\end{tabular}
\end{table*}

\section{Experimental Setup and Results}

Here, we include more details and discussion on experimental setup, datasets and results.

\subsection{Experimental Setup}
\label{app:setup}
Here, we provide more details on experimental setup.
\label{subsec:appendix_setup}
We perform grid search for hyperparameter tuning.
For IRM, IRM-MMD and IRM-ACDM, we search the iteration number to plug in the IRM penalty term ($K_{IRM}$) in $200,400,600$ and IRM penalty weight $\alpha$ in $\{1,10,...,10^8\}$.
For MMD, ACDM, IRM-MMD, IRM-ACDM, we search CDM penalty weight $\beta$ in $\{1,10,...,10^5\}$.
For ACDM and IRM-ACDM, we set the number of steps we train the discriminator $D$ in each iteration to $10$.
For EIIL, we do the same hyperparameter tuning on the IRM penalty weight ($\alpha$), the iteration to add IRM ($K_{IRM}$) as we do for IRM, IRM-ACDM and IRM-MMD methods. We train the soft environment weight $q(E|X,Y)$ for $\{10,100,1000,10000\}$ steps.

\subsection{Datasets}
\label{app:data}
Here, we present more details about the datasets.

\noindent\textbf{CMNIST.} CMNIST is introduced by~\citep{arjovsky2019invariant}.
We can see the two training domains of CMNIST are similar to each other in terms of both $P(C|Y,E)$ and $P(Y|E)$ in Table~\ref{tab:colored_mnist}.
This means CMNIST does not cover the case of strong $\Lambda$ spurious, since the spurious correlations, color--domain and domain--class, are not strong.


\begin{table*}[tbh!]
\caption{Description of CMNIST}
\label{tab:colored_mnist}
\centering
\small
\begin{tabular}{|c|c|c|c|c|c|}
\hline
$E$                      & $P(Y=1|E)$             & $Y$   & $P(C=G|Y,E)$ & $P(C=B|Y,E)$ & $P(C=R|Y,E)$ \\ \hline
\multirow{2}{*}{$E=1$} & \multirow{2}{*}{0.5} & $Y=1$ & $0.9$          &  $0.0$        & $0.1$        \\ \cline{3-6} 
                       &                      & $Y=0$ & $0.1$         & $0.0$       & $0.9$        \\ \hline
\multirow{2}{*}{$E=2$} & \multirow{2}{*}{0.5} & $Y=1$ & $0.8$        & $0.0$         & $0.2$       \\ \cline{3-6} 
                       &                      & $Y=0$ & $0.2$         & $0.0$        & $0.8$        \\ \hline
\multirow{2}{*}{$E=3$} & \multirow{2}{*}{0.5} & $Y=1$ & 0.1          & 0.0         & 0.9        \\ \cline{3-6} 
                       &                      & $Y=0$ & 0.9          & 0.0        & 0.1        \\ \hline
\end{tabular}
\end{table*}

\begin{figure*}[tbh!]
\centering
\includegraphics[width=\textwidth]{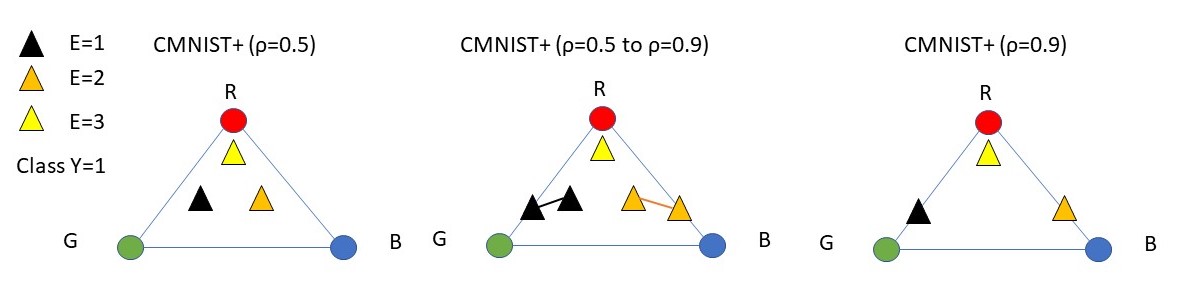}
\caption{We visualize CMNIST+ with $\rho\in[0.5,0.9]$ in terms of $P(C|Y=1,E)$. Each large triangle represents the space of $P(C|Y,E)$. Each small triangle shows the values of $P(C=c|Y=1,E=e), c\in\{R,G,B\}, e\in\{1,2,3\}$. } \label{fig:cminst_plus_vis}
\end{figure*}

\noindent\textbf{CMNIST+.} We visualize the CMNIST+ dataset with different values of $\rho$ in Fig.~\ref{fig:cminst_plus_vis}. In addition, we provide a detailed simulation recipe of CMNIST+ and compare it with that of CMNIST. This would also show that CMNIST+ is in accordance with the causal graph in Fig.~\ref{fig:causal_graph_cmnist_plus}.
\begin{enumerate}
\item We decide the true label $Y^*$ (without noise) of each instance by its original digit label (0-9).
\item We randomly split the data into test and training.
\item We generate noisy labels $Y$ by randomly flipping them with 25\% probability. This means $Y^*\rightarrow Y$.
\item We assign the training instances to the two training domains based on the noisy label $Y$ and $P(Y|E)$ in Table 1. This step introduces correlations between $Y$ and $E$. In each training environment, we further randomly split the data into training and validation.


\item Given $P(C|Y,E)$ in Table~\ref{tab:colored_mnist_plus}, the noisy label and the domain label, we assign color to each instance. This step introduces correlations among $C$, $Y$ and $E$.

\end{enumerate}

Note that there is a difference in what causal relationships mean in traditional causal inference and in OOD prediction. In OOD prediction, the definition of causal relationships is different from a traditional one. Traditionally, $S \rightarrow Y$ means the generation of $Y$ is (partially) determined by $S$. It does not necessarily mean $P(Y|S)$ remains the same across domains. However, in OOD prediction, we say there exists a causal relationship $S \rightarrow Y$ iff $P(Y|S)$ is the same across different domains.
We also know in the original MNIST dataset, there exists invariant causal relationship $S\rightarrow Y^*$.
This implies that, from the data generating process of CMNIST+, we confirm that (1) there exist invariant relationships $S\rightarrow Y^* \rightarrow Y$, (2) there are spurious correlations among $C$, $Y$ and $E$.
With these two conclusions, we can claim that the causal graph in Fig.~\ref{fig:causal_graph_cmnist_plus} is in accordance with CMNIST+ in the OOD prediction problem.

\noindent\textbf{Creating Strong Triangle Spuriousness with Two Colors.}
It is possible to setup strong triangle spuriousness with two colors for binary classification with two training domains.
Here, we use the two colors: red (R) and green (G).
To show it is possible, we use an example with the following setup: $P(Y=1|E=1)=0.9$, $P(Y=1|E=2)=0.1$. Let’s say for $E=1$, we set $P(C=G|Y=1,E=1)=0.9$, $P(C=G|Y=0,E=1)=0.1$. Then, we can set $P(C=G|Y=1,E=2)=0.1$, $P(C=G|Y=0,E=2)=0.9$ for $E=2$. This setup makes strong correlations among the color, the class label and the domain label. Thus, it is possible to create strong triangle spuriousness with just two colors.
Our concern with such datasets is that even if strong triangle spuriousness exists in training domains, it is a challenge to create test domains that are diverse enough from the training ones.
Following the aforementioned setup, for the test domain $E=3$, if we set $P(Y=1|E=3)=0.5$, $P(C=G|Y=1,E=3)=0.5$ and $P(C=G|Y=0,E=3)=0.5$, it would be right in the middle of the two training domains.
Unfortunately in this setting, even if IRM fails, it can be difficult to observe it with the test accuracy.
This is because a model perfectly fits the color features can reach $0.5$ test accuracy. This leads to smaller differences between models fitting causal features and those with spurious features.
Thus, it becomes more challenging to judge whether a model fails in practice, which explains why we use three colors in CMNIST+.

\end{document}